\documentclass[12pt,onecolumn]{article}
\usepackage[latin1]{inputenc}
\usepackage{amsmath,amsfonts,amssymb,graphicx,authblk,placeins,cite}
\usepackage{comment}
\usepackage{float}

\begin{document}

\pagenumbering{gobble}

\title{How deep is deep enough ?\\
  Quantifying class separability in the\\ hidden layers of deep neural networks
}

\author[1,2]{Achim Schilling}
\author[2,3]{Claus Metzner}
\author[2]{Jonas Rietsch}
\author[3]{Richard Gerum}
\author[2]{Holger Schulze}
\author[1,2]{Patrick Krauss}
\affil[1]{\small Cognitive Computational Neuroscience Group at the Chair of English Philology and Linguistics, Department of English and American Studies, Friedrich-Alexander University Erlangen (FAU), Germany}
\affil[2]{Experimental Otolaryngology, Neuroscience Lab, University Hospital Erlangen, Friedrich-Alexander University Erlangen (FAU), Germany}
\affil[3]{Biophysics Group, Department of Physics, Friedrich-Alexander University Erlangen (FAU), Germany}

\maketitle

\newpage
\noindent\textbf{Corresponding author:} \\
Dr. Patrick Krauss  \\
Neuroscience Group \\
Experimental Otolaryngology \\
Friedrich-Alexander University of Erlangen-N\"urnberg (FAU) \\
Waldstrasse 1 \\
91054 Erlangen, Germany \\
Phone:  +49 9131 85 43853 \\
E-Mail: patrick.krauss@uk-erlangen.de \\ \\


\newpage
\begin{abstract}\large
Deep neural networks typically outperform more traditional machine learning models in their ability to classify complex data, and yet is not clear how the individual hidden layers of a deep network contribute to the overall classification performance. We thus introduce a Generalized Discrimination Value (GDV) that measures, in a non-invasive manner, how well different data classes separate in each given network layer. The GDV can be used for the automatic tuning of hyper-parameters, such as the width profile and the total depth of a network. Moreover, the layer-dependent GDV(L) provides new insights into the data transformations that self-organize during training: In the case of multi-layer perceptrons trained with error backpropagation, we find that classification of highly complex data sets requires a temporal {\em reduction} of class separability, marked by a characteristic 'energy barrier' in the initial part of the GDV(L) curve. Even more surprisingly, for a given data set, the GDV(L) is running through a fixed 'master curve', independently from the total number of network layers. Furthermore, applying the GDV to Deep Belief Networks reveals that also unsupervised training with the Contrastive Divergence method can systematically increase class separability over tens of layers, even though the system does not  'know' the desired class labels. These results indicate that the GDV may become a useful tool to open the black box of deep learning.
\end{abstract}


\newpage
\section*{Introduction}

Recently, Artificial Intelligence (AI) and Machine Learning (ML) have moved into more and more domains such as medicine, science, finance, engineering, or even entertainment, and are about to become ubiquitous in 21st century life. Especially in the field of Deep Learning, the pace of progress has been extraordinary by any measure, and Deep Neural Networks (DNNs) are performing extremely well in a vast number of applications such as image classification, or natural language processing \cite{lecun2015deep}. In combination with reinforcement learning, the networks are becoming proficient in playing video games \cite{mnih2015human}, or, by playing against themselves, are reaching super-human levels in complex board games, such as Go \cite{silver2016mastering, silver2017mastering}.

At the same time, AI and ML are facing several crises. In particular, many results published in ML are difficult to reproduce, since these results seem to depend sensitively on small details of the training conditions, which are often not well documented \cite{hutson2018artificial}. Also, the optimal parameter settings in ML projects are usually found by mere trial-and-error, a state of affairs that has been called the 'alchemy' problem \cite{sculley2018winner}. Both the reproducibility and alchemy problem are related to the fundamental opacity of Deep Learning Networks: We do not currently have a theoretical understanding of how internal data representations emerge in the different layers of a DNN during the training process \cite{szegedy2013intriguing, lin2017does}. We also cannot predict how the performance of a DNN will depend on its many hyper-parameters, such as the number of layers, or the sizes of each layer, given a specific task. DNNs therefore still must be considered as 'black boxes' \cite{voosen2017ai}, and to change this status would require, besides theoretical work, new tools for analysis and optimization.

A first 'glimpse into the black box' of DNNs is provided by data visualization techniques, such as t-distributed stochastic neighbor embedding (t-SNE) \cite{maaten2008visualizing} or multi-dimensional scaling (MDS) \cite{torgerson1952multidimensional}, which project the high-dimensional activation vectors of each network layer onto points in two (or three) spatial dimensions. By color-coding each projected data point of a training-data set according to its desired output class label, the representation of the data in a given network layer can be visualized as a set of point clusters. In principle, the apparent compactness and mutual overlap of these point clusters permits a qualitative assessment of how well the different classes separate. However, apart from the problem that the resulting low-dimensional projections can be highly dependent on the detailed parameter settings of the visualization method (in particular with t-SNE \cite{wattenberg2016use}), it is often difficult to compare the degree of separability for two given projections that stem from the same layer but different training histories, and impossible to do so for representations drawn from layers of different dimensionality. 

For this reason, we provide in this work a new measure of class separability that quantifies and objectifies the intuitive notions of compactness and mutual overlap of the point clusters, however without projecting the data and without requiring any free parameters. The measure, called the General Discrimination Value (GDV), is defined as the difference between the mean intra-cluster variability and the mean inter-cluster separation, computed on a set of labeled, z-scored vectors in n-dimensional space. The GDV is zero for data points with randomly shuffled labels, and minus one in the case of perfect class separability. Furthermore, it is invariant under a global shift or scaling of the data vectors, as well as invariant under a permutation of the neuron indices. Due to proper normalization, the GDV makes it possible for the first time to quantitatively compare the degree of class separability between two layers that contain different numbers of neurons (dimensionality invariance), between networks trained with a different number of examples for each label (class-size invariance), or even between networks trained for tasks of different complexity (class-number invariance). These features make the GDV an ideal tool to open the black box of deep learning.

In principle, class separability can also be quantified with the traditional classification accuracy. However, while the GDV works directly with the continuous, distributed data representation of a network layer, the accuracy, being defined as the fraction of correct classifications, always depends on the 'one-hot' representation of a classifier output. Therefore, measuring the accuracy at some hidden layer L in the network requires to add a classification layer after the point of interest, and then to train this classification layer for the desired labels. Unfortunately, the resulting accuracy is then no longer a property of the first L network layers alone, but a property of the total, distorted system. By contrast, the GDV is a 'non-invasive' measure that does not distort the network in any way.

The GDV can provide new insights about deep learning, both in supervised and unsupervised settings. In supervised learning, the most common task is classification, where continuous input vectors are mapped onto discrete output labels. Since the input vectors in DNNs are typically high-dimensional, whereas there are only few possible output labels, classification is necessarily accompanied by data compression, and this is usually reflected in a monotonically decreasing number of neurons in subsequent layers of the neural network. Thus, while parts of the input information are eliminated in each processing step, it is critical for the DNN to discard only irrelevant information that does not contribute to creating the desired output label. This process of gradually eliminating irrelevant details of the data while retaining its relevant aspects can be visualized geometrically by the point clusters that are associated with each class: The centers of the clusters (the 'prototypical' realizations of each class) represent relevant information, whereas the widths of the clusters (the intra-class variability) can be considered irrelevant for the classification task. 

After supervised training of a DNN, the average intra-class variability should therefore diminish in successive network layers, while the average inter-cluster separation should remain constant or even increase. By design, both would yield to a decrease of the GDV. Indeed, we find a monotonous decrease of the GDV with the layer index, at least for data sets of relatively low complexity (such as MNIST \cite{LeCun1998, lecun2010mnist} or fashion-MNIST \cite{xiao2017fashion}). However, our study indicates that more complex data sets (such as CIFAR-10 \cite{krizhevsky2014cifar} or Caltech-101 \cite{fei2007learning}) cannot immediately be separated into distinct classes, but first require certain 'preparatory' transformations that do not change or even temporally increase the GDV. Only after this initial phase, the GDV begins to fall quickly, and eventually saturates or continues to fall more slowly. This layer-dependent GDV(L) curve can be used to optimize the hyperparameters of the network, such as the total depth of the network and the widths of the individual layers. 

We also analyze unsupervised learning with our new tool. In particular, we investigate Deep Belief Networks (DBNs) \cite{hinton2009deep}, trained layer-wise with the Contrastive Divergence method \cite{hinton2002training} on the MNIST data set. Remarkably, although these models do not have the predefined objective to separate input data into distinct classes, and do not even 'know' the set of possible class labels, the GDV is consistently decreasing for tens of layers, both for constant and decreasing layer widths. Since the GDV is falling faster for a less complex subset of the data, this finding indicates that Contrastive Divergence training can detect and separate clusters of similar inputs even in non-labeled data.


\FloatBarrier
\newpage
\section*{Results}

\vspace{1cm}
\subsection*{Validation of GDV with artificial data}

\begin{figure}[ht!]
\centering
\includegraphics[]{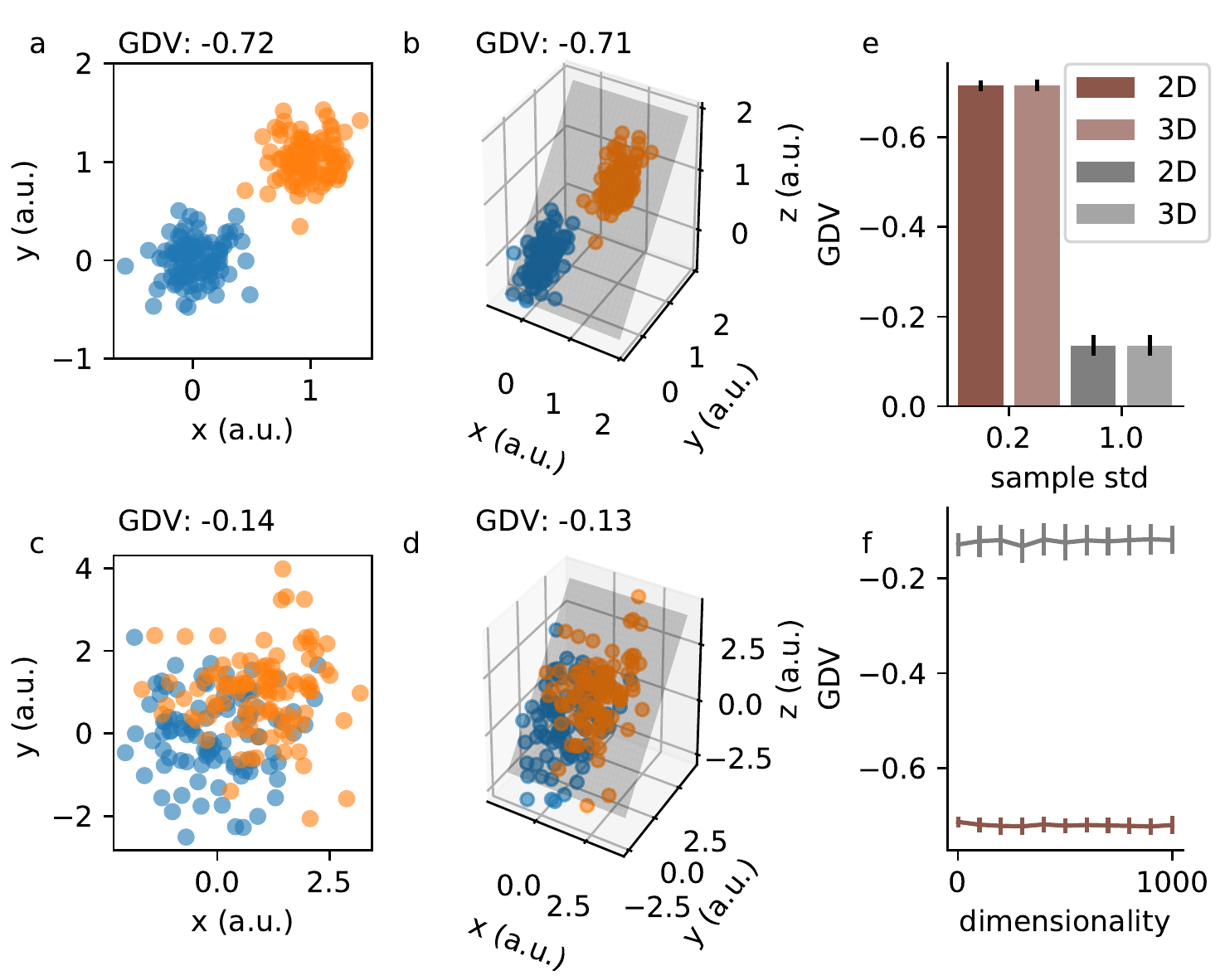}
\caption{
Demonstration of GDV using artificial data. 
(a): Two easily separable clusters result in a GDV of -0.72. (b): Embedding the two-dimensional data into  three-dimensional space, by mapping points $(x_k,y_k)$ onto $(x_k,y_k,y_k)$, does not change the GDV. 
(c,d): Two overlapping Gaussian distributions result in a GDV of -0.14.
(e) Summary of the test cases (a-d).
(f) Embedding the data from (a) and (b) into increasingly high-dimensional spaces leaves the GDV invariant.
}
\label{fig1}
\end{figure}

\newsavebox{\covmata}
\savebox{\covmata}{$\left(\begin{smallmatrix}0.04&0\\0&0.04\end{smallmatrix}\right)$}
\newsavebox{\covmatb}
\savebox{\covmatb}{$\left(\begin{smallmatrix}1&0\\0&1\end{smallmatrix}\right)$}
The GDV is defined as the normalized difference between the average intra-cluster variability and the average inter-cluster separation, computed on a set of labeled, z-scored vectors in n-dimensional space (Eq.\ref{GDVEq}). The features of this quantity can be demonstrated using artificial data (Fig.\ref{fig1}): Two well-separated clusters, generated from distinct two-dimensional Gaussian distributions with $\boldsymbol{\mu_1}=(0,0)^T$, $\boldsymbol{\mu_2}=(1,1)^T$, and $\boldsymbol{\Sigma_1}=\boldsymbol{\Sigma_2}=$\usebox{\covmata}, lead to a GDV of $-0.72$ (Fig. \ref{fig1}a). By contrast, two overlapping clusters, generated from distributions with $\boldsymbol{\mu_3}=(0,0)^T$, $\boldsymbol{\mu_4}=(1,1)^T$, and $\boldsymbol{\Sigma_3}=\boldsymbol{\Sigma_4}=$\usebox{\covmatb}, lead to a GDV of $-0.14$ (Fig. \ref{fig1}c). 
Since the GDV is based on the Euclidean distance between data points, it is invariant with respect to a rigid translation of the data, as well as invariant to a permutation of the neuron indices. Furthermore, the GDV is only sensitive to the effective data subspace and does not change when the data is embedded into arbitrary higher-dimensional spaces. For example, embedding the two-dimensional Gaussian data from above into three (Fig. \ref{fig1}b,d,e) or more dimensions (Fig. \ref{fig1}f),   has no effect on the GDV.

\vspace{1cm}
\subsection*{Multi-layer perceptrons (MLPs), trained with\\ error backpropagation}

In this work, the GDV is used as a tool to quantify cluster separability in different layers $L$ of neural networks. We start with multi-layer perceptrons (MLPs) \cite{ruck1990feature} of either constant or decreasing layer widths (cf. Methods), which are trained using error backpropagation on various data sets (MNIST \cite{LeCun1998, lecun2010mnist}, fashion-MNIST \cite{xiao2017fashion}, and CIFAR-10 \cite{krizhevsky2014cifar}). Subsequently, we compute the GDV for each network layer in each model, using the corresponding test data sets. Additionally, we visualize the clusters of data points in selected layers (input layer, and hidden layers 2, 9, 15) using multi-dimensional scaling (MDS).

\subsubsection*{$GDV(L)$ curve consists of three regimes\\ and depends on data complexity}

\begin{figure}[ht!]
\centering
\includegraphics[]{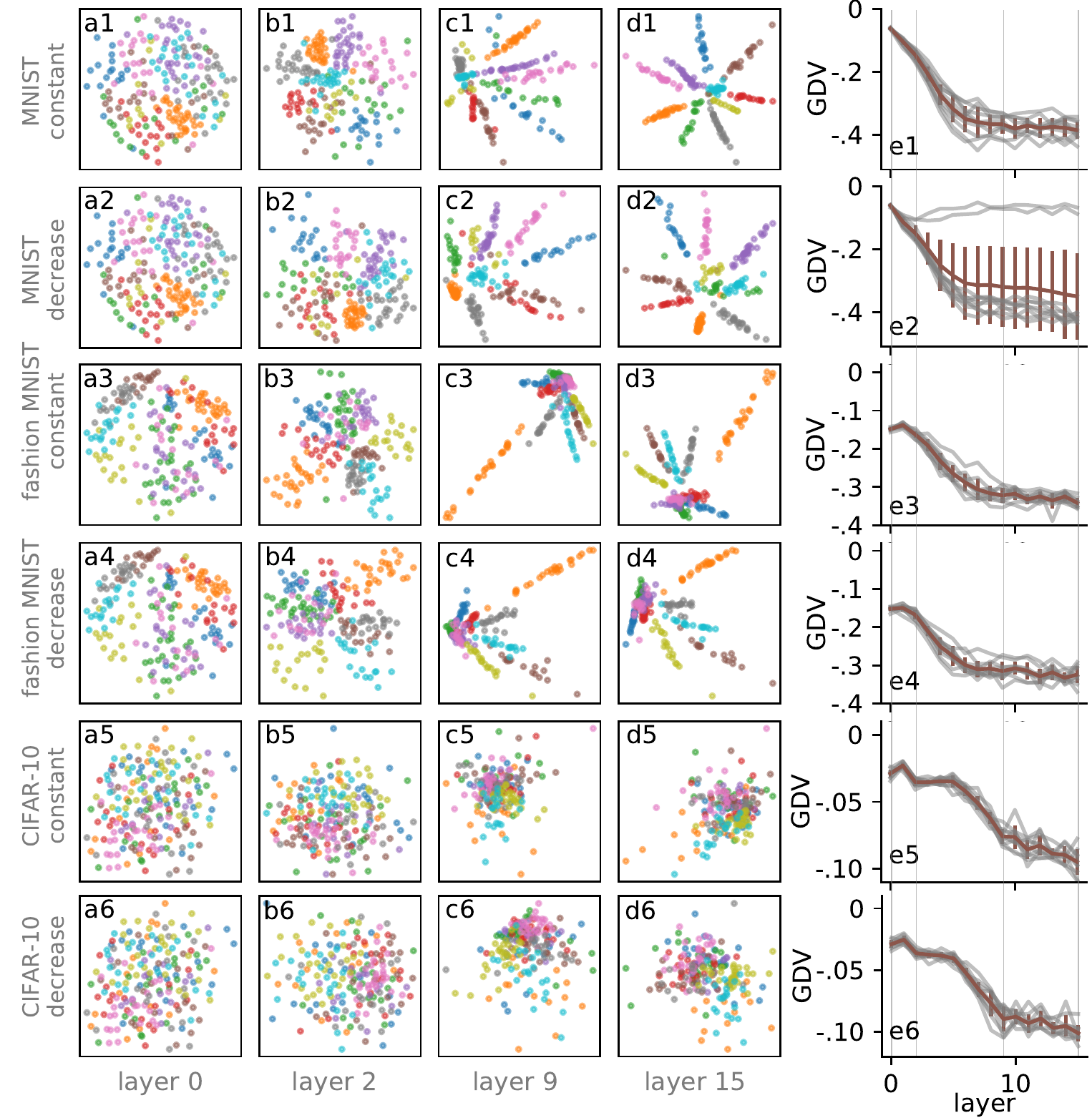}
\caption{Layer-dependent class separation in multi-layer perceptrons, trained on the MNIST (rows 1-2), fashion-MNIST (rows 3-4) and CIFAR-10 (rows 5-6) data sets, using either constant (even rows) or decreasing (odd rows) layer widths. Left four columns show MDS projections of the test data sets in selected layers (0, 2, 9 and 15), with colors corresponding to data set labels. Right column shows the GDV as function of layer index L. The curve GDV(L) consists of an optional initial phase, a phase of rapid decay, and a final phase, where the GDV remains constant or continues to fall slowly. Fine gray lines in the GDV(L) plots depict the layers for which MDS projections (a-d) were computed. Clusters become clearly separable as soon as the GDV(L) enters the final phase. For the CIFAR-10 data set, both  GDV(L) and MDS results indicate that the given 15 layers are not sufficient for a perfect class separation.
}
\label{fig2}
\end{figure}

When an MLP is used as a classifier, the distinct data classes are supposed to separate well in the final output layer. However, it is not clear how class separability develops over the hidden layers of the network. We therefore compute the GDV as a function of layer index $L$. We find that the curve $GDV(L)$ consists of up to three characteristic regimes: An (optional) initial regime, where the GDV may occasionally increase (Fig. $\,$\ref{fig2}e, 3-6), a regime of rapid decay (Fig. $\,$\ref{fig2}e, 1-6), and a final regime where the GDV either saturates (Fig. $\,$\ref{fig2}e, 1-4) or continues to decrease more slowly (Fig. $\,$\ref{fig2}e, 5-6).
The number $n_i$ of network layers belonging to the initial regime correlates with the complexity of the data set, which increases from MNIST ($n_i=0$), over fashion-MNIST ($n_i=2$), to CIFAR-10 ($n_i=5$). The minimum GDV at the transition between the rapid decay and the final regime also correlates with data complexity (about -0.4 for MNIST, -0.3 for fashion-MNIST, and -0.1 for CIFAR-10). Interestingly, however, the number $n_r$ of network layers belonging to the rapid decay regime does not seem to depend on data complexity ($n_r=5$ in all cases). Finally, we note that in the case of the MNIST dataset, the ADAM optimizer failed to find a good minimum in two out of ten independent network trainings (Fig. $\,$\ref{fig2} e2).

\subsubsection*{GDV(L) is consistent with multi-dimensional scaling analyis}

For the MNIST and fashion-MNIST  data sets (Fig. $\,$\ref{fig2}, first four rows), the monotonous decrease of the GDV curve is reflected in a gradual demixing and compactification of the clusters in the MDS projections. For the CIFAR-10 data set (Fig. $\,$\ref{fig2}, last two rows), strongly overlapping clusters in the MDS projection of input layer 0 indicate a larger data complexity. Consequently, even in layer 15, where the GDV is still only about -0.1, the separation of the clusters in the MDS plot is not significantly better than in the input layer.  

\subsubsection*{GDV(L) reveals optimum model hyper-parameters}

The shape of the $GDV(L)$ curve can be used to determine optimal model hyper-parameters: As rule of thumb, the last layer of rapid GDV decrease can be considered as the optimal network depth for classification. Moreover, GDV(L) also indicates the optimum layer widths: If, for a given data set, the curves $GDV(L)$ are identical for networks with constant and decreasing layer width (as is the case for all data sets in Fig. $\,$\ref{fig2}) one would opt for the model with fewer parameters.

\subsubsection*{GDV(L) is independent from network depth}

Since error backpropagation works from the last layer towards the input layer, one might expect that in shallow and deep networks different representations of the input emerge, resulting in different GDV values at corresponding layers L of these networks. To test this hypothesis, we trained multi-layer perceptrons with a total number of 3, 7, 11 and 15 layers on the CIFAR-10 data set (Fig. 3). Surprisingly, the GDV(L) all follow a common 'master curve', suggesting that for each data set there exists an optimal hierarchy of increasingly complex features, and that the highest level of complexity that can be processed is limited by the network depth. Moreover, the GDV(L) of the training data set (blue) is consistently smaller (better separation of data classes) than that of the test data set (gray) in all network layers.

\subsubsection*{GDV correlates with test accuracy}

In machine learning, performance of a classification task is usually quantified by the test accuracy. In order to test if classification accuracy correlates with class separability in the output layer, we have computed both quantities for multi-layer perceptrons with 15 layers of equal width, trained on the MNIST (dark blue), the fashion-MNIST (light blue), and the CIFAR-10 (cyan) data sets (Fig. 4). The increasing complexity of these three data sets is reflected in relative values of the assymptotic GDV (a,c). Furthermore, we find indeed a monotoneous relation between the GDV in the final network layer and the classical accuracy (b,d). Note that a comparison between GDV and accuracy is not possible for the hidden layers, since computing the accuracy requires to insert and train a fully connected 'one-hot' output layer at the point of interest, which affects the original representations in an unknown way. By contrast, the GDV is a 'non-invasive' measure that can be directly calculated from the input-driven activations at any hidden layer.

\vspace{1cm}
\subsection*{ResNet50 pre-trained on ImageNet\\ and tested with Caltech-101 data set}

We also compute the layer-specific GDV for the ResNet50 network \cite{he2016deep}, trained on the ImageNet data set \cite{deng2009imagenet}. The ResNet50 is a large state-of-the-art model with an architecture that is not linear, but has numerous parallel paths. The GDV is evaluated for all of these paths (Fig. \ref{fig5}, q), but we mainly focus on the 'add' layers of the network, where several paths are converging.

\subsubsection*{GDV can test separability of classes different\\ from training data set}

In addition, we now take advantage of the GDV's generality and use a data set for testing, in which not only the patterns, but also the classes are different from the training data set. Specifically, we choose a subset of the Caltech-101 data set \cite{fei2007learning} for testing. As a result, we find that GDV(L) never reaches the phase of rapid decay for all of the 175 layers (Fig. \ref{fig5}, q), which is not surprising considering both, the complexity of and the mismatch between train and test data. 

\subsubsection*{GDV strongly affected by final softmax layer}

As already mentioned above, computing the accuracy necessarily disturbs the learned representations in all network layers before the point of interest, and this is particularly true if train and test data classes do not match. Here, this disturbance is reflected in a drastic GDV drop in second last (fully connected) layer and a final increase in the last (softmax) layer (orange markers in Fig. \ref{fig5}, q).

\vspace{1cm}
\subsection*{Deep Belief Networks (DBNs), trained layer-wise\\ with contrastive divergence}

Finally, we apply the GDV to a generative model trained in an unsupervised manner. In particular, we use a Deep Belief Network (DBNs), trained layer-wise with contrastive divergence on the MNIST data set.

\subsubsection*{GDV reaches a minimum even with unsupervised learning}

Even though the network has no classification objective and receives no information about class labels, the GDV(L) decreases consistently for about 30 layers (Fig. \ref{fig6}a), both for constant (dark red) and decreasing (gray) layer widths. After this point, the GDV(L) is slowly increasing again. This behaviour is almost identical for networks with constant and decreasing layer widths. 

Moreover, as in the examples of supervised learning above, the GDV(L) curve correlates with data complexity: the GDV at layer zero is smaller and decreases more steeply (Fig. \ref{fig6}b, red) for the less complex, i.e. easy to discriminate, digits (0, 1, 6). In contrast, the GDV starts at a larger value and decreases less steeply (Fig. \ref{fig6}b, cyan) for more complex digits (4, 7, 9) with rather similar shapes. 

\subsubsection*{GDV minimum is confirmed by dreamed prototype patterns}

This decrease is also reflected in the prototypical inputs for each digit and each image (Fig. \ref{fig6}c): prototype patterns are first blurry (e.g. in layers 1 and 5) and then become increasingly clear up to about layer 30, where the GDV reaches its minimum (Fig. \ref{fig6}a). Beyond layer 30, blurriness is increasing again.


\newpage
\section*{Discussion}

In this work, we have introduced the GDV as a new, parameter-free measure of class separability in deep neural networks. In contrast to the traditional accuracy, the GDV can be evaluated 'non-invasively' in any hidden layer of the network, i.e. it does not require adding a classification layer after the point of interest. Moreover, the set of test data patterns, and even the target classes used to compute the GDV can be chosen different from the training data, thus allowing to asses the generality of the learned features. Finally, the GDV is  independent from dimensionality, so that class separability in two layers with different widths can be directly compared.

We have applied the GDV to a large variety of neural network types, such as MLPs \cite{ruck1990feature}, ResNet50 \cite{he2016deep}, and DBNs \cite{hinton2009deep} (main manuscript), as well as ConvNets \cite{dumoulin2016guide}, LSTMs \cite{hochreiter1997long}, VGG19 \cite{simonyan2014very}, Xception \cite{chollet2017xception}, InceptionV3 \cite{szegedy2016rethinking}, NASNet Mobile \cite{zoph2018learning}, and stacked auto-encoders \cite{bengio2007greedy} (supplemental material). These models were trained on classification of several data sets from the image domain, such as MNIST \cite{LeCun1998, lecun2010mnist}, fashion-MNIST \cite{xiao2017fashion}, CIFAR-10 \cite{krizhevsky2014cifar}, ImageNet \cite{deng2009imagenet}, Caltech-101 \cite{fei2007learning}, and also on sentiment classification of a natural language data set, the Internet Movie Database (IMDb) \cite{maas2011learning}. 

By computing the GDV for all layers of these neural networks, we have analyzed for the first time in a quantitative way how class separability changes along the processing chain of deep learning systems. We have demonstrated that the curve $GDV(L)$ provides novel insights - and also stimulates new research questions - regarding the complexity of the data, their sequential transformation within the network, and regarding the effect of hyper-parameter changes: 

The GDV in the input layer $L=0$ quantifies the intrinsic degree of clustering that is already present in the input data, before any processing by the network. For example, the intrinsic clustering of the fashion-MNIST dataset ($GDV(0)\approx -0.15$) is significantly smaller than that of the CIFAR-10 dataset ($GDV(0)\approx -0.03$), and this is  reflected in the corresponding MDS projections (Fig. \ref{fig2}, a3 versus a5). Similar differences in $GDV(0)$ are seen within the MNIST data set for subsets of digits with different complexity (Fig. \ref{fig2}c). Thus, the GDV(0) measures the complexity, or difficulty, of a labeled data set with respect to classification.

Complex data sets can cause the $GDV(L)$ to remain constant or even to {\em increase} over the initial layers of a neural network, suggesting that certain preparatory transformations are required that temporarily merge and recombine, rather than separate data classes. In future work, it would be interesting to investigate if the shape of this initial 'energy barrier' in $GDV(L)$ can be controlled by using artificial data sets with known properties. 

In the case of deep multi-layer perceptrons, arguably the network type with the simplest and most regular structure, the $GDV(L)$ curve can be divided into up to three characteristic phases, which exist independently from the data set: After the optional initial 'barrier' phase, the $GDV(L)$ always reaches a phase of rapid decrease, before it eventually saturates or continues to fall at a much slower rate. Although our preliminary results already indicate a clear correlation of these $GDV(L)$ curves with data complexity, it may be worthwhile to explore in detail which aspects of a data set control the widths and slopes of the three phases.

Strikingly, in multi-layer perceptrons trained on the complex CIFAR-10 data set, the $GDV(L)$ runs through a fixed 'master curve' that is independent from the total network depth. This seems to indicate that there exists only one optimal sequence of transformations that eventually renders the data separable into distinct classes. Why networks of different depth cannot use different strategies to disentangle the data classes remains to be investigated.

The $GDV$ can also serve as an objective function for an automatic optimization of hyper-parameters, such as the width of every individual layer in a deep network. Indeed, in the case of the multi-layer perceptrons, we have already demonstrated that almost identical $GDV(L)$ curves result, no matter if the layer widths remain constant or moderately decrease towards the output layer. Future work will reveal which are the minimal layer widths that can be used for a given classification task without significant loss of performance.

In a similar way, the asymptotic behavior of the $GDV(L)$ curve in a trained network indicates whether class separability could be further increased by adding more layers to the network. For example, this seems to be the case with the multi-layer perceptrons trained on the CIFAR-10 data set.  

A further remarkable finding was that class separability is improving systematically over many layers in a Deep Belief Network with constant layer width, even though it was 'trained' in an unsupervised, label-free manner using the Contrastive Divergence method. A similar effect has been pointed out before by Bengio et al. \cite{bengio2007greedy, bengio2009learning} and may be related to a hidden tendency of auto-encoder-like layers to compress data into effectively lower-dimensional representations, while the number of neurons in the input and output layer remains formally constant. This data compression, in turn, can result in a decreasing GDV. 

Another unresolved question with the Deep Belief Network is why the $GDV(L)$ curve is eventually increasing again after about layer 30. We speculate that the network, having no information about the user-defined classes, is eventually 'over-generalizing' and starts to re-merge classes that were well separated before. This hypothesis is corroborated by the observation that the $GDV(L)$ curve shows no final increase when the network is only confronted with subsets of the data.

Summing up, the GDV may serve as a universal tool to 'open the black box of deep learning' and to move a small step from 'alchemy' towards 'chemistry' of AI.


\FloatBarrier
\newpage
\section*{Methods}

\subsection*{Generalized discrimination value (GDV)}
In a previous paper we introduced the concept of the discrimination value, which quantifies the separability of point clusters in high-dimensional spaces \cite{krauss2018statistical}: The more compact and mutually disjoint the considered clusters are, the smaller the discrimination value becomes. Here we generalize this concept in terms of invariance with respect to scale, dimensionality and number of different labels.
For this purpose, we consider $N$ points $\mathbf{x_{n=1..N}}=(x_{n,1},\cdots,x_{n,D})$, distributed within $D$-dimensional space. A label $l_n$ assigns each point to one of $L$ distinct classes $C_{l=1..L}$. In order to become invariant against scaling and translation, each dimension is separately z-scored and, for later convenience, multiplied with $\frac{1}{2}$:
\begin{align}
s_{n,d}=\frac{1}{2}\cdot\frac{x_{n,d}-\mu_d}{\sigma_d}.
\end{align}
Here, $\mu_d=\frac{1}{N}\sum_{n=1}^{N}x_{n,d}\;$ denotes the mean, and $\sigma_d=\sqrt{\frac{1}{N}\sum_{n=1}^{N}(x_{n,d}-\mu_d)^2}$ the standard deviation of dimension $d$.
Based on the re-scaled data points $\mathbf{s_n}=(s_{n,1},\cdots,s_{n,D})$, we calculate the {\em mean intra-class distances}  
\begin{align}
\bar{d}(C_l)=\frac{2}{N_l (N_l\!-\!1)}\sum_{i=1}^{N_l-1}\sum_{j=i+1}^{N_l}{d(\textbf{s}_{i}^{(l)},\textbf{s}_{j}^{(l)})},
\end{align}
and the {\em mean inter-class distances}
\begin{align}
\bar{d}(C_l,C_m)=\frac{1}{N_l  N_m}\sum_{i=1}^{N_l}\sum_{j=1}^{N_m}{d(\textbf{s}_{i}^{(l)},\textbf{s}_{j}^{(m)})}.
\end{align}
Here, $N_k$ is the number of points in class $k$, and $\textbf{s}_{i}^{(k)}$ is the $i^{th}$ point of class $k$.
The quantity $d(\textbf{a},\textbf{b})$ is the distance between $\textbf{a}$ and $\textbf{b}$ in a suitable metric (see below).
Finally, the generalized discrimination value $\Delta$ is calculated from the mean intra-class and inter-class distances  as follows:
\begin{align}
\Delta=\frac{1}{\sqrt{D}}\left[\frac{1}{L}\sum_{l=1}^L{\bar{d}(C_l)}\;-\;\frac{2}{L(L\!-\!1)}\sum_{l=1}^{L-1}\sum_{m=l+1}^{L}\bar{d}(C_l,C_m)\right].
 \label{GDVEq}
\end{align}

The resulting discrimination value becomes $-1.0$ if two clusters of Gaussian distributed points are located such that the mean inter cluster distance is two times the standard deviation of the clusters.

\subsubsection*{Choice of distance metrics $d(\textbf{a},\textbf{b})$}

In principle, one may use any suitable distance metric $d(\textbf{a},\textbf{b})$ between two D-dimensional points $\textbf{a}$ and $\textbf{b}$ to compute the GDV, such as the Euclidean, Mahalanobis, Manhattan, or Hamming distance. Recently, doubts have been raised about the applicability of the Euclidean distance in high dimensions \cite{aggarwal2001surprising}. However, we perform a z-scoring on each dimension independently and thereby resolve the problem of different scaling in each dimension, which was the main issue with the Euclidean distance identified in the Aggarwal paper \cite{aggarwal2001surprising}.
Furthermore, a paper by Walters-Williams  \cite{walters2010comparative}, which systematically compares several distance metrics, concludes that 'if one does not have any prior knowledge the Euclidean function is usually recommended'. Moreover, since the Euclidean distance, or L2-norm, is frequently used in many successful machine learning applications (e.g. as a loss function), we opted for the L2-norm:
$d(\textbf{a},\textbf{b})=\sqrt{\sum_{d=1}^{D}(a_d-b_d)^2}$.

\subsection*{Computational resources, software and data}

All simulations were implemented in Python 3.6 and performed on a high-performance PC, equipped with an i9e decacore CPU and two Nvidia TitanXp GPUs. For efficient mathematical operations, the following Python libraries were used: NumPy and SciPy \cite{VanDerWalt2011a}, and scikit-learn \cite{pedregosa2011scikit} for mathematical operations, Matplotlib \cite{Pedregosa2012} and Pylustrator \cite{Gerum2018} for the visualization of the data. Furthermore, the neural networks were implemented using Keras \cite{chollet2015keras} and TensorFlow \cite{Abadi2016}.

\subsection*{Multi-layer perceptrons}

Two different multi-layer perceptron (MLP) architectures of either constant or decreasing layer width have been used for training. Both networks consisted of an input layer and 15 fully connected dense layers. They where trained on the MNIST \cite{LeCun1998, lecun2010mnist}, fashion-MNIST \cite{xiao2017fashion} and CIFAR-10 \cite{krizhevsky2014cifar} data set with error backpropagation, using the ADAM optimizer \cite{kingma2014adam}. The input layer consisted of $28 \times 28 = 784$ neurons (for MNIST and fashion-MNIST), or $32 \times 32 \times 3 = 3072$ neurons (for CIFAR-10), respectively. In case of the constant layer width MLP, each hidden layer consisted of 256 neurons, whereas the decreasing width MLP's first hidden layer consisted of 256 neurons, the second one of 246, and so on. The last hidden layer's width was 116.

\subsection*{Deep belief networks}

Deep Belief Networks (DBNs) are a class of autoencoders that have been introduced by Geoffrey Hinton \cite{hinton2009deep}. They consist of Restricted Boltzmann Machines (RBMs) \cite{hinton1983optimal,hinton2006reducing,larochelle2008classification}, stacked in such a way that each RBM's hidden layer serves as the visible layer for the next RBM. Thus, DBNs can be trained greedily and without supervision, one layer at a time, by applying methods such as Contrastive Divergence learning \cite{carreira2005contrastive,hinton2012practical, hinton2006fast}. As a result, a hierarchy of feature detectors is emerging, which can later be used for classification tasks. Moreover, being a generative model, DBNs can be used to probabilistically reconstruct the input.

\subsubsection*{Boltzmann neurons}

RBMs are based on Boltzmann neurons \cite{hinton1983optimal}. The total input $z_i(t)$ of neuron $i$ at time $t$ is calculated as:
\begin{equation}
	\mathrm{z_i(\textit{t})} = b_i + \sum\limits_{j=1}^{N} w_{ij}\;y_j(t-1)\, ,
\end{equation} 
where $y_j(t-1)$ is the binary state of neuron $j$ at time $t-1$,  $w_{ij}$ is the weight from neuron $j$ to neuron $i$, and $b_i$ is the bias of neuron $i$. The probability $p_i(t)$ of neuron $i$ to be in state $y_i(t)=1$ is given by: 
\begin{equation}
	p_i(t) = \sigma(z_i(t)),
\end{equation}
where $\sigma(x)$ is the logistic function
\begin{equation}
	\sigma(x) = \frac{1}{1\;+\;e^{-x}}.
\end{equation}

\subsubsection*{Visualizing learned prototype digits in the DBN} 

Based on the DBN with constant width, we re-construct the prototypical input patterns $PIP(C,L)$ for each digit class $C$ and for each network layer $L$, adapting a technique from deep dreaming \cite{le2008representational}: We choose a test data image $K$ from class $C$, apply it to the network input, and compute the average representation of this image in layer $L$. Next, we perform a winner-takes-all sparsification, by setting the ten percent most active neurons of this layer to one and all others to zero. This sparsified activity is then reversely propagated through all layers, down to the $28\times28$ input matrix, where it results in an image-specific input pattern $IP(K)$. The prototypical input pattern $PIP(C,L)$ for digit class $C$ is computed by averaging the $IP(K)$ over all test images $K$ from this class $C$. Results of this procedure are summarized in Fig. \ref{fig6}.

\FloatBarrier


\newpage

\newpage

\newpage
\begin{figure}[ht!]
\centering
\includegraphics[]{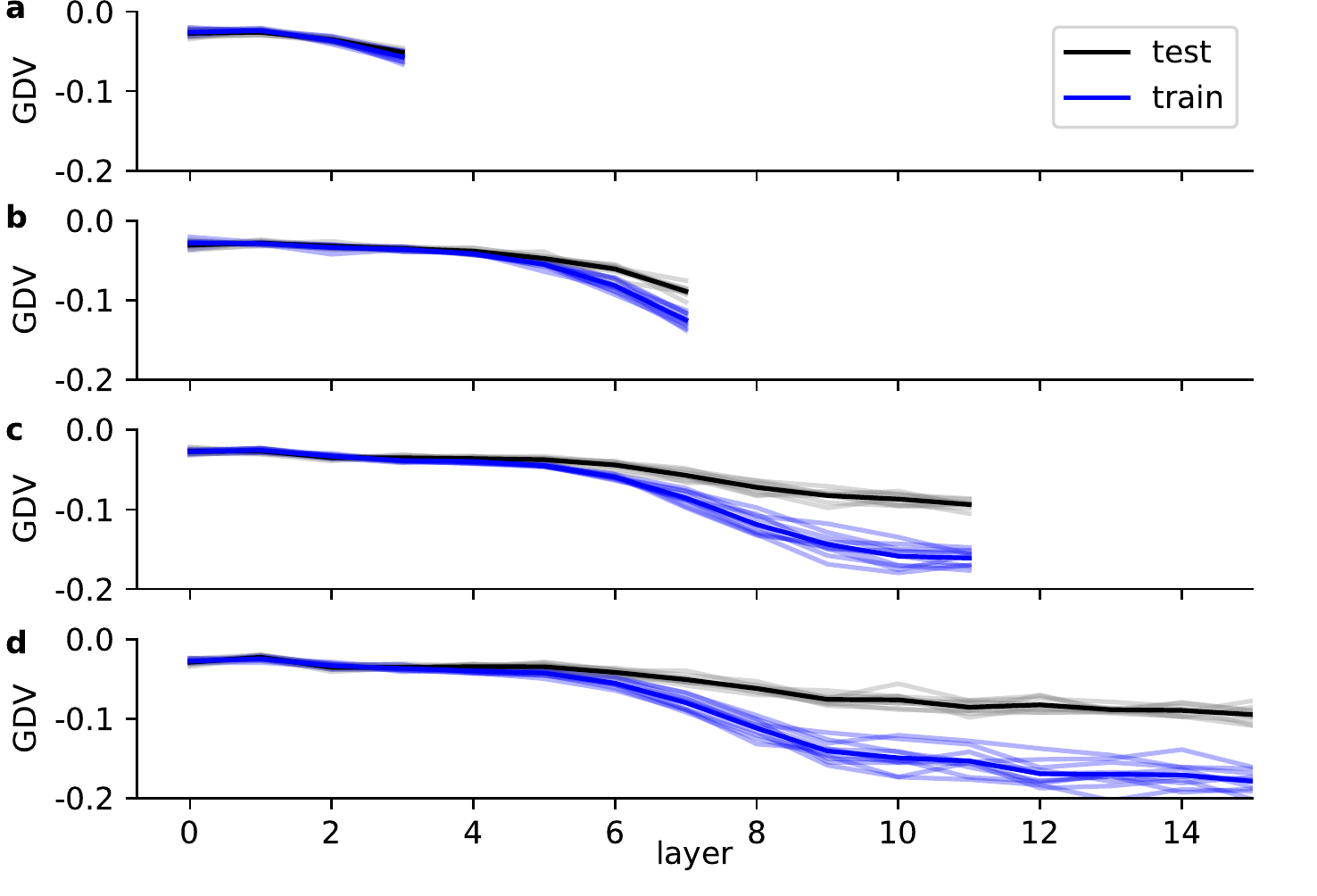}
\caption{Layer-dependent class separation in multi-layer perceptrons of different network depth (rows a-d correspond to 3, 7, 11 and 15 layers), trained on the CIFAR-10 data set. The black curves show the GDV(L) for the test data set, the blue curves for the training data set. Remarkably, all curves seem to follow the same course, independent of the network depth.}
\label{fig3}
\end{figure}

\newpage
\begin{figure}[ht!]
\centering
\includegraphics[]{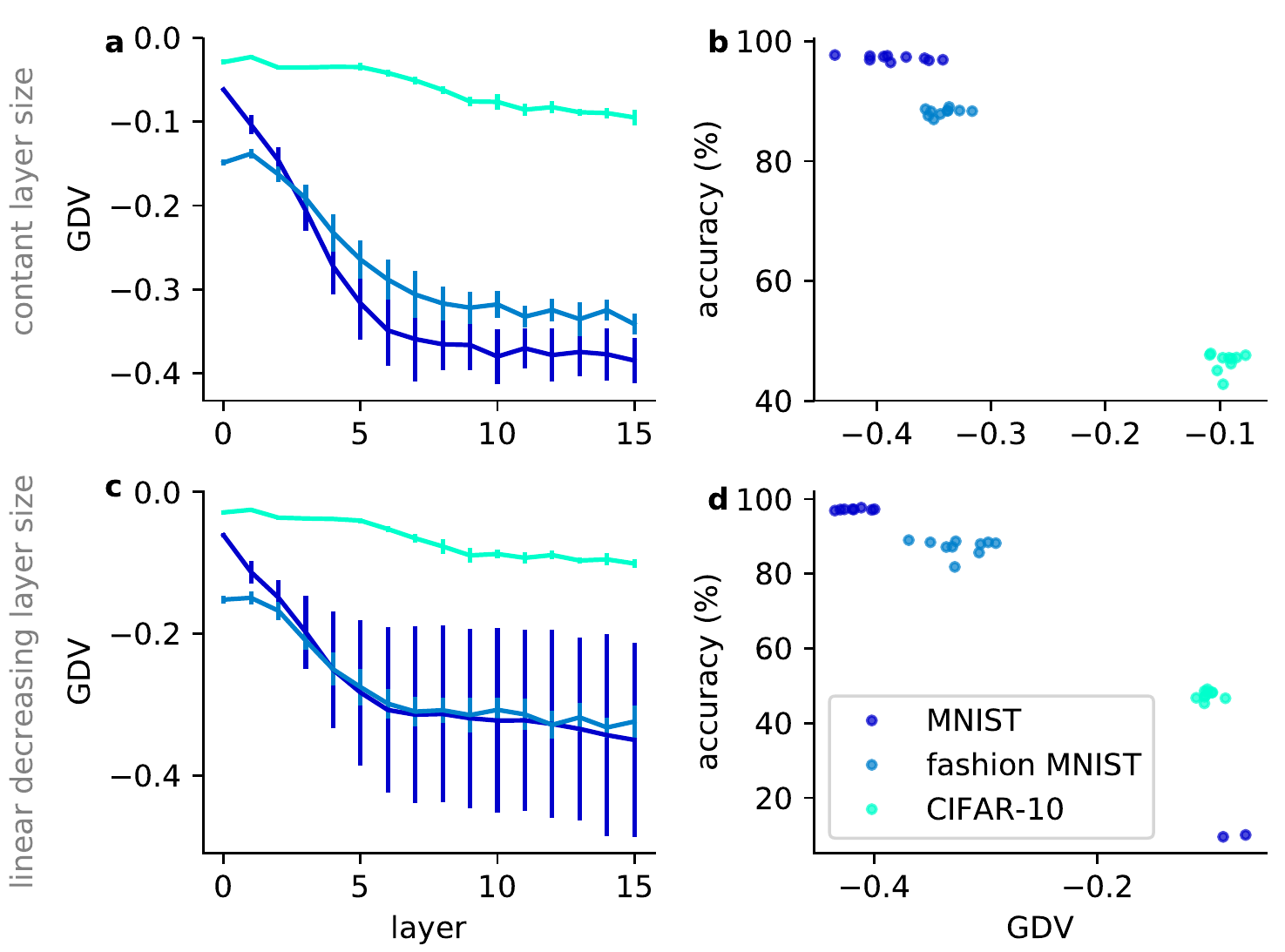}
\caption{Multi-layer perceptrons with 15 layers of equal width, trained on the MNIST (dark blue), the fashion-MNIST (light blue), and the CIFAR-10 (cyan) data sets. The GDV(L) curves in the left panel demonstrate again how data complexity affects class separablity. The right panel shows that there is a monotoneous relation between the GDV in the final network layer and the classical accuracy. Note that multiple data points per data set in b and d correspondd to multiple runs of the network training algorithm.}
\label{fig4}
\end{figure}

\newpage
\begin{figure}[ht!]
\centering
\includegraphics[]{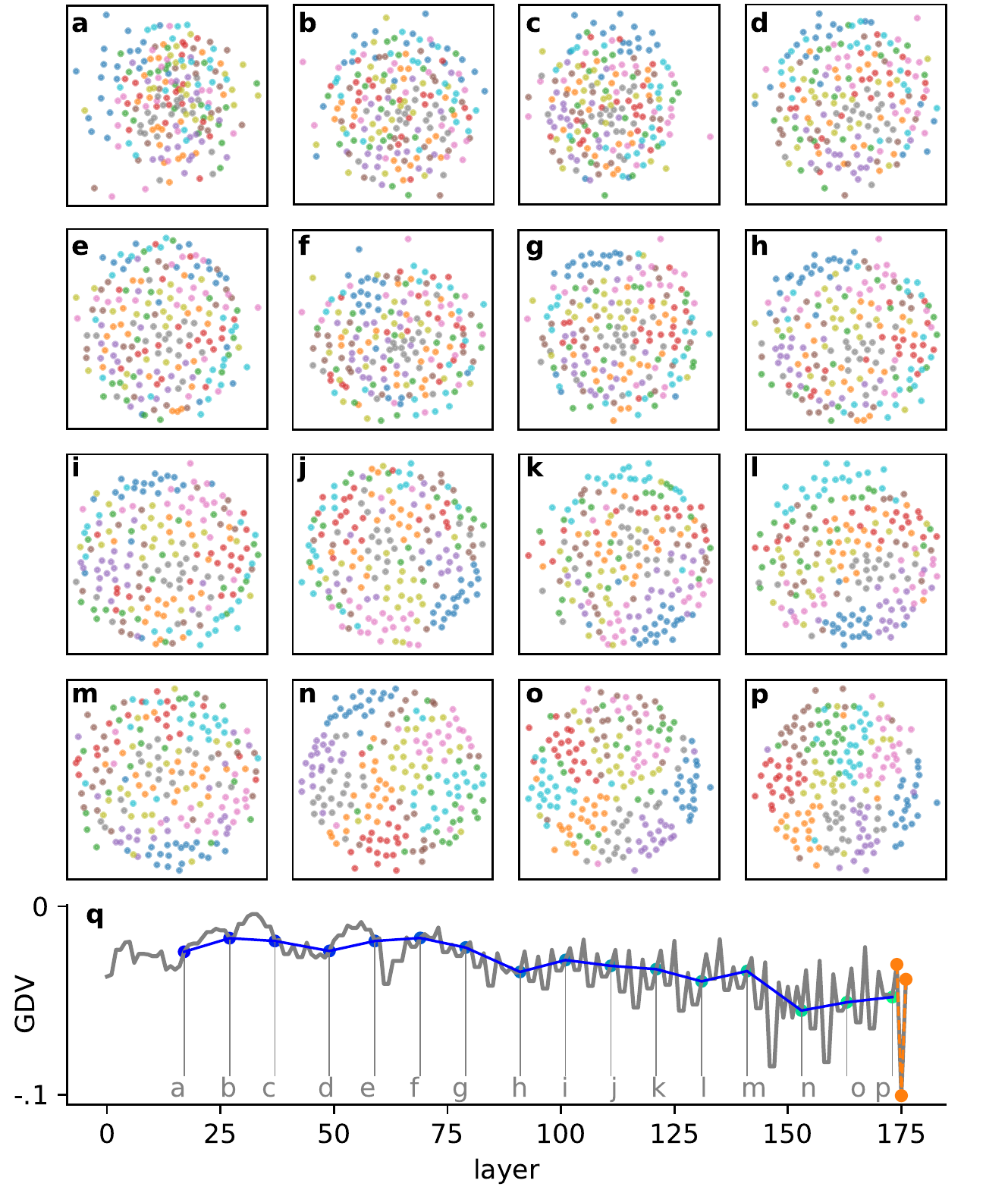}
\caption{Layer-dependent class separation in a ResNet50 network, trained on the ImageNet data set, but evaluated with a subset of the Caltech-101 data set. The MDS projections (a-p) correspond to the 'add' layers of the network, which are indicated by the the fine gray lines in the GDV(L) plot (q, bottom row). The GDV drops sharply in the second last layer to a new global minimum, and increases again in the final output layer (orange). The drop to the global minimum is due to the convergence of spatially separated feature channels in a fully connected dense layer. The final increase of GDV reflects the disturbance caused by the softmax layer.}
\label{fig5}
\end{figure}

\newpage
\begin{figure}[ht!]
	\centering
	\includegraphics[width=0.7\linewidth]{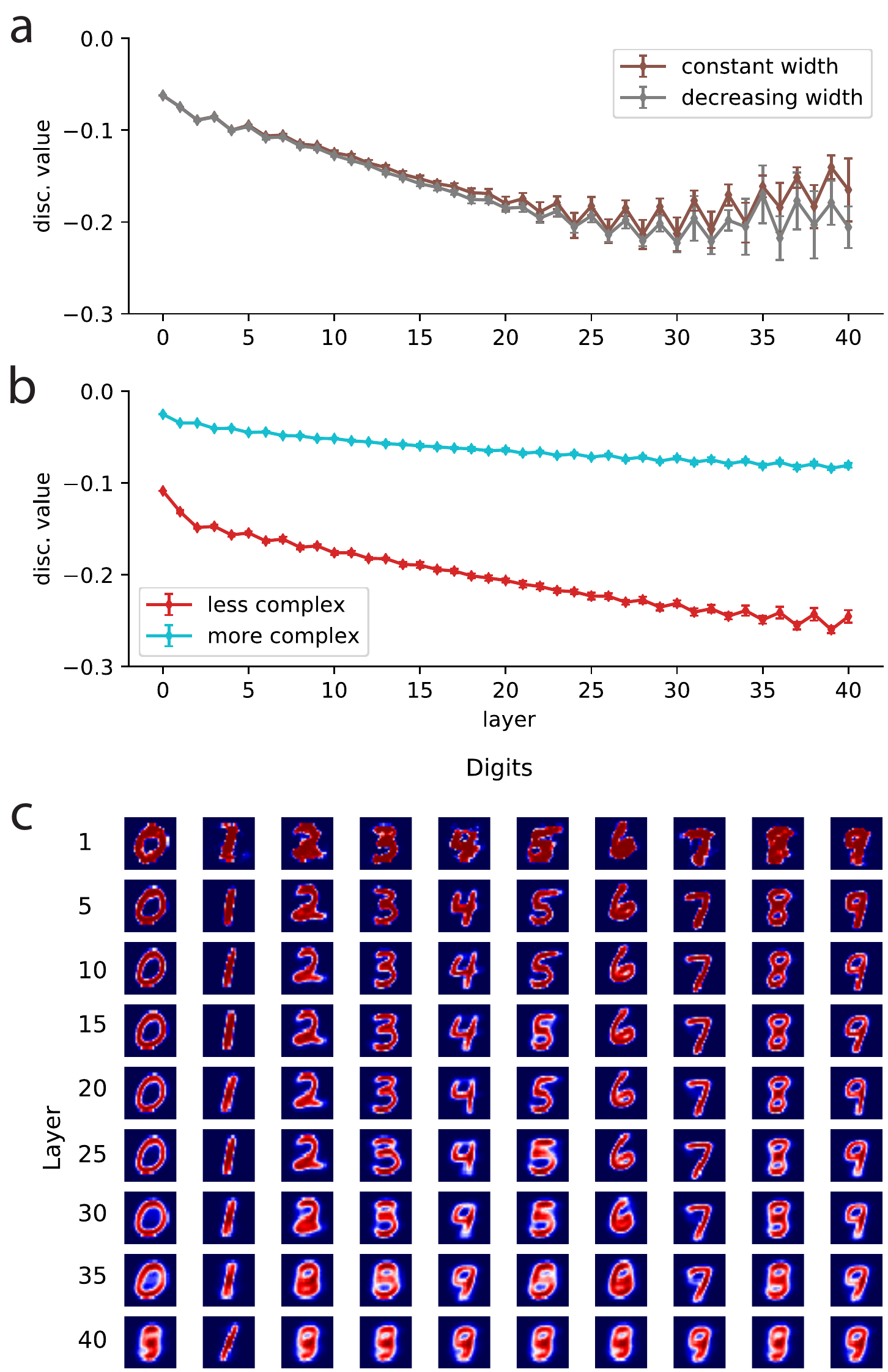}
	\caption{Layer-dependent class separation in a Deep Believe Network, trained unsupervised with the Contrastive Divergence method on the MNIST data set. Even though the network has no classification objective and receives no information about class labels, the GDV(L) decreases consistently for about 30 layers (a), both for contant (dark red) and decreasing (gray) layer widths. The GDV at layer zero is smaller and decreases more steeply (b, red) for the less complex, i.e. easy to discriminate, digits (0, 1, 6). In contrast, the GDV starts at a larger value and decreases less steeply (b, cyan) for more complex digits (4, 7, 9), which have very similar shapes. This decrease is also reflected in the prototypical inputs for each digit and each image (c): prototype patterns are first blurry (e.g. in layers 1 and 5) and then become increasingly clear up to about layer 30, where the GDV reaches its minimum (a). Beyond layer 30, blurriness is increasing again.} 
	\label{fig6}
\end{figure}


\FloatBarrier
\newpage
\section*{Additional Information}

\subsection*{Data availability statement} 
All data will be made available online.

\subsection*{Acknowledgments}
This work was supported by the Deutsche Forschungsgemeinschaft (DFG, grants SCHU1272/12-1 and ME1260/11-1), and the Interdisciplinary Center for Clinical Research Erlangen (IZKF, ELAN-17-12-27-1-Schilling) at the University Hospital of the University of Erlangen-Nuremberg. The authors are grateful for the donation of two Titan Xp GPUs by the NVIDIA
Corporation.

\subsection*{Author contributions}
PK, CM and AS designed the study. AS, CM, RG, JR and PK performed the computer simulations. AS, PK and CM developed the theoretical approach. AS, PK, RG and CM analyzed the neural networks. CM, PK, AS and HS wrote the paper. All authors read and approved the final manuscript.

\subsection*{Competing interests}
The authors declare no competing financial interests.


\newpage
\section*{SUPPLEMENTAL MATERIAL}

\subsection*{Effect of random transformations on the GDV}

The General Discrimination Value is a function, $GDV=f(U,L)$, which maps a given data set U and an associated set L of labels onto a scalar value. Here, a data set U is a list of $M$ data points $U=\left\{\vec{u}_1,\vec{u}_2,\ldots,\vec{u}_M \right\}$, each represented by a $N$-dimensional vector $\vec{u}_m=(u_{m1},u_{m2},\ldots,u_{mN})$. The label set L is a list of integers $L=\left\{l_1,l_2,\ldots,l_M \right\}$, which assigns a specific class $l_m$ to each data point $\vec{u}_m$. Usually, it is assumed that the number of distinct classes $K$ is smaller than the number of data points $M$.

\vspace{0.2cm}\noindent
Since the GDV depends only on (averages over) the Euklidean distances between pairs of data points, it is trivially invariant with respect to
permutations of the $N$ coordinates, and with respect to global shifts of all data points. Due to the z-scoring, the GDV is also invariant with respect to a linear scaling of the coordinates. Furthermore, as we have demonstrated above, is remains also invariant when the data is embedded into higher-dimensional spaces.

\begin{figure}[ht!]
\centering
\includegraphics[width=12cm]{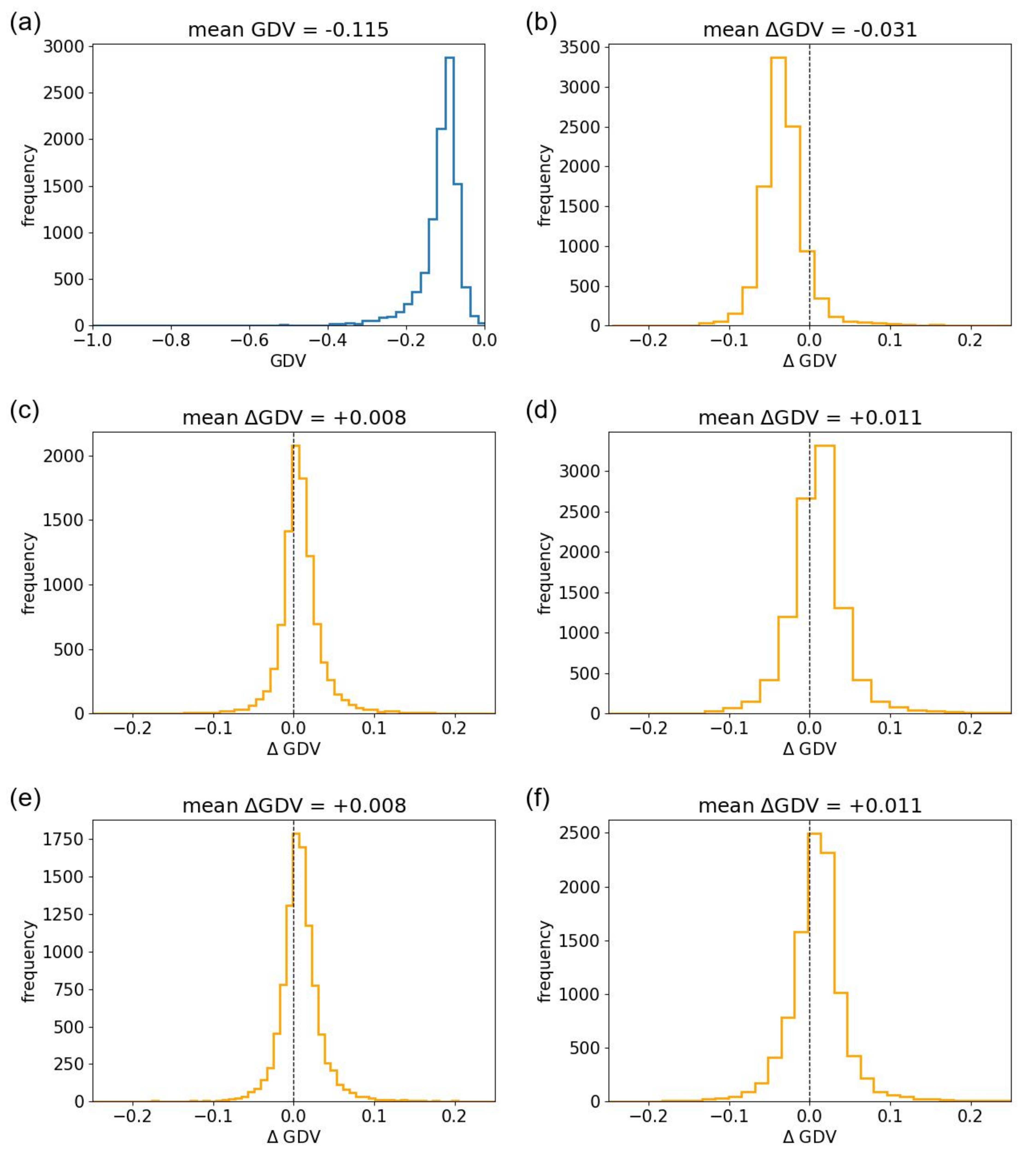}
\caption{
\small
Empirical distribution $p(GDV)$ of the General Discrimination Value GDV over random Gaussian data sets (for details, see main text), and distributions $p(\Delta GDV)$ of the GDV changes after applying various transformations.
{\bf(a)} Distribution $p(GDV)$ over an ensemble of $10^4$ random data sets. 
{\bf(b)} Distribution $p(\Delta GDV)$ of GDV changes after scaling all dimensions by a factor of 10 and subsequently applying the logistic function. 
{\bf(c)} Distribution $p(\Delta GDV)$ of GDV changes  after applying a random transformation matrix with elements drawn independently from a uniform distribution in the range $\left[ -10,+10 \right]$.
{\bf(d)} Same as case c, however with additional application of the logistic function. 
{\bf(e,f)} Same as cases c and d, however using non-square shaped random matrices that doubled the number of dimensions.
Note that for cases c-f the resulting mean change of GDV is positive and of the approximate magnitude $\Delta GDV\approx 0.01$, which is about 10 percent of the mean GDV itself. Only in case b, the mean change of GDV is negative and its magnitude corresponds to about 30 percent of the mean GDV.  
}
\label{figSM1}
\end{figure}

\vspace{0.2cm}\noindent
Here, we additionally investigate the effect of various more complex transformations on the GDV. For this purpose, we generate an ensemble of $10^4$ artificial data sets (each represented as 'point clusters' in multidimensional space) with widely varying properties. In particular, each data set is assigned a different number $N$ of dimensions (drawn randomly between 2 and 10, with equal probabilities), a different number $K$ of classes (drawn randomly between 2 and 10, with equal probabilities), and a different number $S$ of points per class (drawn randomly between 1 and 100, with equal probabilities). The geometrical center point $\vec{\mu}_k$ of each class $k$ is drawn randomly from a uniform distribution within the $N$-dimensional unit-cube $\left[0,1\right]^N$. All data points belonging to a given class $k$ are distributed around their center point $\vec{\mu}_k$ according to a Gaussian distribution, where each dimension $n$ can have a different standard deviation $\sigma_n$ (drawn randomly from a uniform distribution between 0 and 1). We have computed the GDV for each of these artificial data sets and find values that fluctuate approximately between -0.4 and 0 (for the distribution, see Fig.\ref{figSM1}(a)),  with an average GDV of -0.115.

\vspace{0.2cm}\noindent
Next, we apply random linear transformations to our artificial data sets. Each such transformation is described by an $N\times N$ transformation matrix $\mathbf{A}$, which is generated by drawing the matrix elements $A_{ij}$ randomly and independently from a uniform distribution between -10 and +10. For maximal variability, a new matrix $\mathbf{A}$ is drawn independently for each data set. We compute the GDV before and after the linear transformation, and then consider the change $\Delta GDV = GDV_{trans}-GDV_{before}$. Over the complete ensemble, the change $\Delta GDV$ fluctuates approximately between -0.1 and 0.1, with a positive mean of +0.008 (for the distribution, see Fig.\ref{figSM1}(c)). The magnitude of this net positive shift is less than 10 percent of the mean GDV itself. We conclude that a random linear transformation of the data can both improve and degrade the separability of classes, but the latter is slightly more probable.

\vspace{0.2cm}\noindent
Within a given layer $j$ of a neural network, the $N$ coordinates of a data point are represented by the activations of $N$ corresponding neurons. Assuming that the subsequent layer $j\!+\!1$ of the network has the same number $N$ of neurons, the linear transformation from above can be realized by random (untrained) neural weights between these two layers. However, a neuron in typical artificial neural networks also applies a non-linear sigmoidal function to the weighted sum $s$ of its inputs, often using the logistic function $y=1/(1+e^{-s})$. We have therefore investigated the effect on the GDV when, after applying the random linear transformation, each coordinate is passed through a logistic function. We find that as a result of the non-linearity, the distribution of the change $\Delta GDV$ is slightly broadened (Fig.\ref{figSM1}(d)), and the mean change is again small (+0.011) and positive. 

\vspace{0.2cm}\noindent
We next consider random linear transformations that lead to a space of different dimensionality, described by non-square-shaped transformation matrices. In particular, we considered transformations from $N$ to $2N$ dimensions, using again matrix elements uniformly distributed between -10 and 10. Both with and without application of the logistic function, we find the same empirical distributions $p(\Delta GDV)$ as in the transformation from $N$ to $N$ dimensions (Figs.\ref{figSM1}(e,f)).

\vspace{0.2cm}\noindent
So far, all considered transformations led to a net positive shift of the GDV, with a small magnitude of about 10 percent of the mean GDV itself. This suggests that, in general, it requires well-optimized matrix elements (neural weights) to actually improve the separation of classes. However, there are interesting exceptions: We have investigated the effect of simply scaling all dimensions by a factor of 10 and then applying the logistic function. As a result (Fig.\ref{figSM1}(b)), we find a significant net negative GDV change of -0.031, which is more than 35 percent of the mean GDV.

\subsection*{GDV for Convolutional Neural Networks trained on MNIST}

We trained a 15 layer ConvNet \cite{lecun2015deep, dumoulin2016guide} on the MNIST data set \cite{LeCun1998, lecun2010mnist} and analyzed the GDV for each layer. The input layer consisted of $28 \times 28 = 784$ neurons. The 15 convolutional layers had 20 filters each. Kernel size was $4 \times 4$ and stride was $1 \times 1$. Padding was set to same and pooling layers were removed.

\begin{figure}[ht!]
	\centering
	\includegraphics[width =\textwidth]{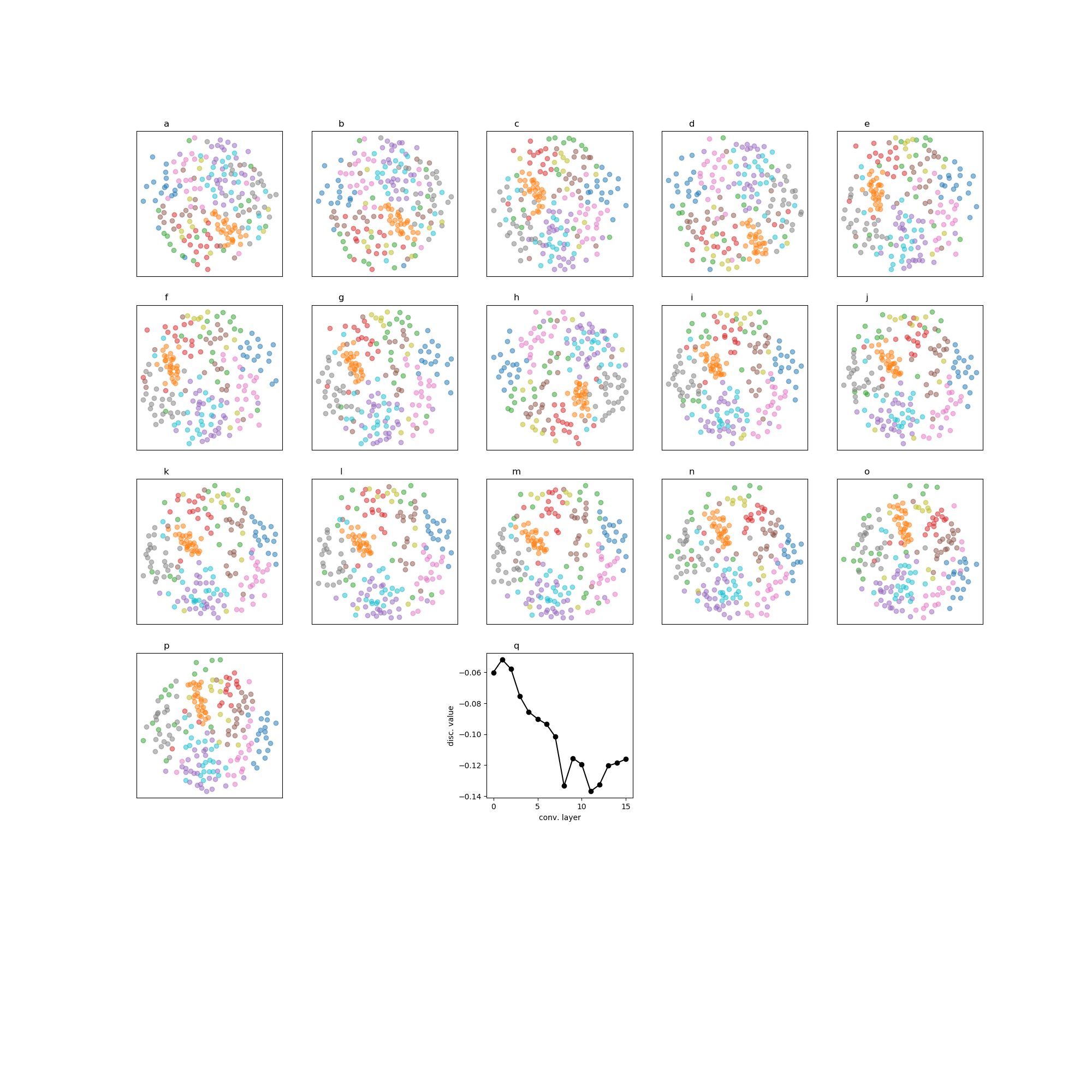}
	\caption{\textbf{GDV(L) of a 15 layer ConvNet trained on MNIST data set.} GDV has 3 phases again.}
	\label{figSM2}
\end{figure}


\subsection*{GDV for LSTMs trained on the IMDb sentiment classification task}

We trained a 6 layer LSTM \cite{hochreiter1997long} on the IMDb sentiment classification task \cite{maas2011learning}. Each LSTM layer consisted of 10 state cells.

\begin{figure}[ht!]
	\centering
	\includegraphics[width = \textwidth]{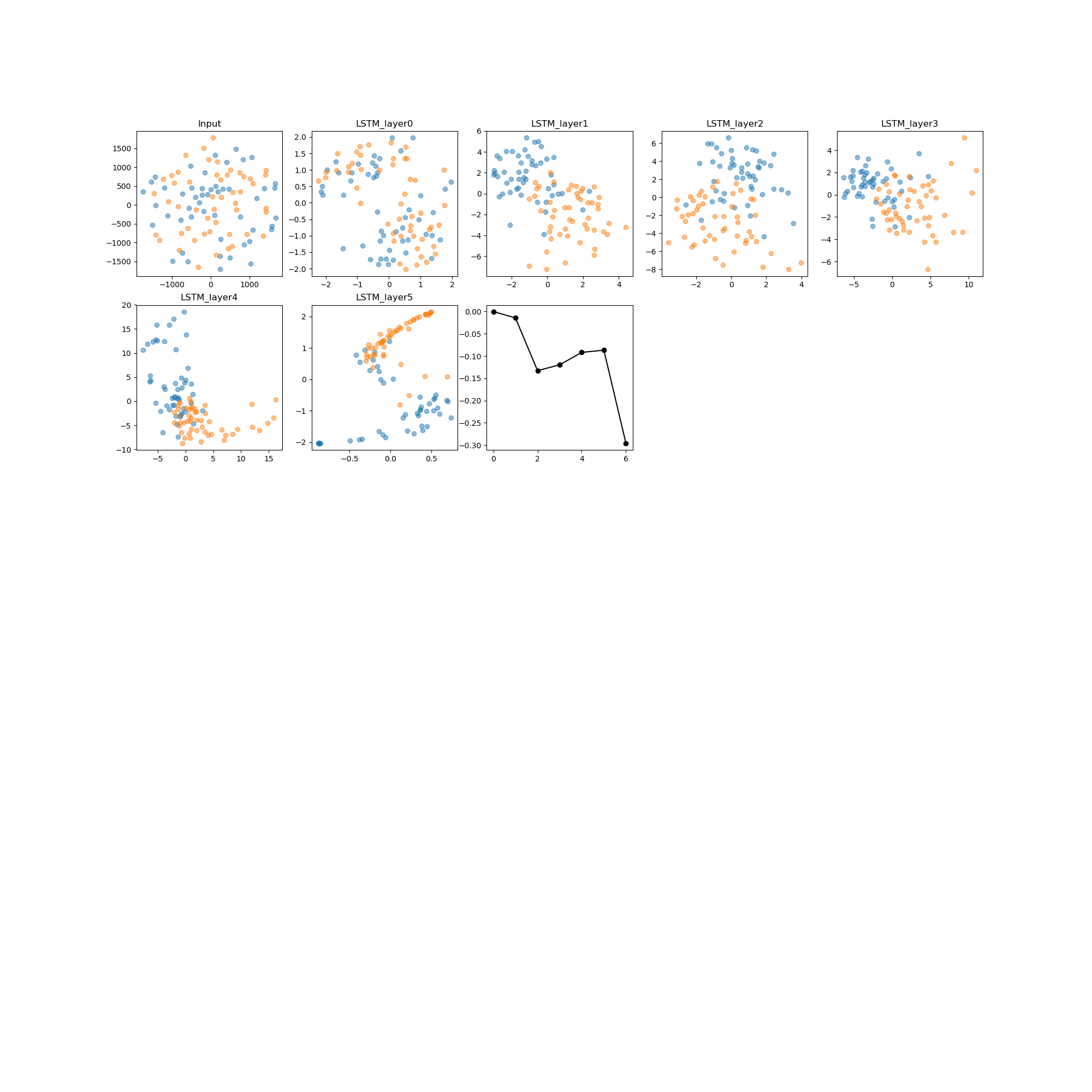}
	\caption{\textbf{GDV for LSTMs trained on the IMDb sentiment task} GDV(L) decreases with depth.}
	\label{figSM3}
\end{figure}


\subsection*{GDV(L) for pre-trained state-of-the-art models and tested with Caltech-101 data set}

\begin{figure}[ht!]
	\centering
	\includegraphics[]{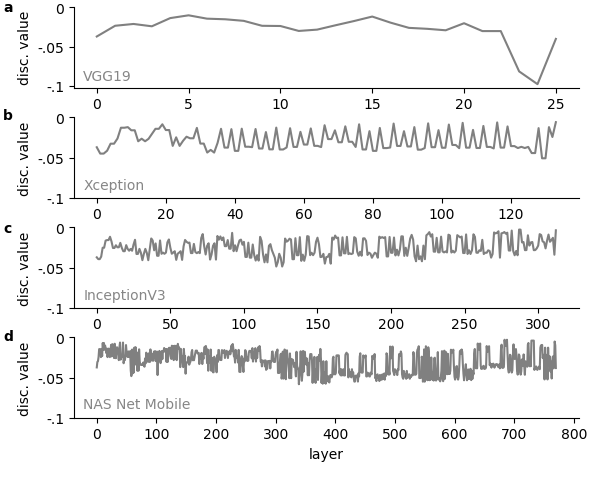}
	\caption{\textbf{GDV(L) for several state-of-the-art networks.} (a) VGG19. (b) Xception. (c) InceptionV3. (d) Nas Net Mobile. Last layer effect potentially because of softmax.}
	\label{figSM4}
\end{figure}

We also compute the layer-specific GDV for the the VGG19 \cite{simonyan2014very}, Xception \cite{chollet2017xception}, InceptionV3 \cite{szegedy2016rethinking}, and NasNet Mobile \cite{zoph2018learning} networks, trained on the ImageNet data set \cite{deng2009imagenet}. These networks are large state-of-the-art models with architectures that are not linear, but have numerous parallel paths. The GDV is evaluated for all of these paths (Fig.\ref{figSM4}) using a sub set of the Caltech-101 data set \cite{fei2007learning}.

\subsubsection*{GDV reveals optimum network depth for compression in stacked auto-encoders}

We trained two different 15 layer stacked auto-encoders \cite{bengio2007greedy}: a constant layer width, and a decreasing layer width auto-encoder. Both models were trained on the MNIST \cite{LeCun1998, lecun2010mnist}, fashion-MNIST \cite{xiao2017fashion}, and CIFAR-10 \cite{krizhevsky2014cifar} data set. Subsequently, the GDV was analyzed for each layer. The input layer consisted of $28 \times 28 = 784$ neurons, or $32 \times 32 \times 3 = 3072$ neurons, respectively The 15 convolutional layers had 20 filters each. In the constant layer width auto-encoder, the hidden layers consisted of 100 neurons, each. In contrast the decreasing layer width network started with 800 neurons in hidden layer 1, 750 neurons in hidden layer 2, and so on, ending with 100 neurons in hidden layer 15.


\begin{figure}[ht!]
	\centering
	\includegraphics[]{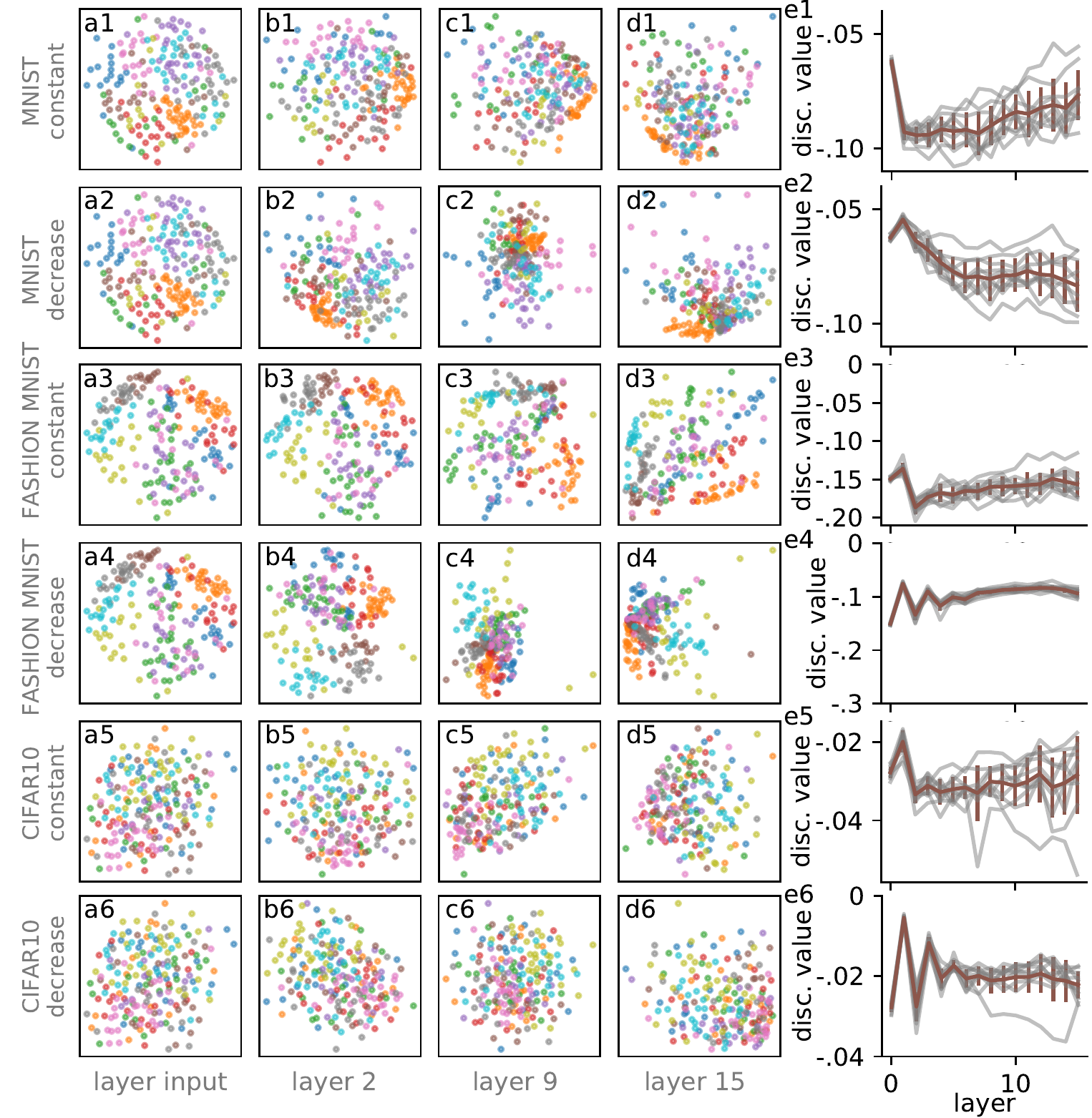}
	\caption{\textbf{GDV(L) for stacked auto-encoders.}}
	\label{figSM5}
\end{figure}


\FloatBarrier

\end{document}